\documentclass[conference]{IEEEtran}
\usepackage{cite}
\usepackage{amsmath,amssymb,amsfonts}
\usepackage{algorithmic}
\usepackage{graphicx}
\usepackage{textcomp}
\usepackage{xcolor}

\begin{document}

\title{A Transfer Learning Framework for Anomaly Detection Using Model of Normality\\

}
\author{\IEEEauthorblockN{Sulaiman Aburakhia}
\IEEEauthorblockA{\textit{Electrical and Computer Engineering} \\
\textit{Western University}\\
London, Canada \\
saburakh@uwo.ca}
\and
\IEEEauthorblockN{Tareq Tayeh}
\IEEEauthorblockA{\textit{Electrical and Computer Engineering} \\
\textit{Western University}\\
London, Canada \\
ttayeh@uwo.ca}
\and
\IEEEauthorblockN{Ryan Myers}
\IEEEauthorblockA{\textit{National Research Council Canada} \\
\text{London, Canada} \\
Ryan.Myers@nrc-cnrc.gc.ca}
\and
\IEEEauthorblockN{Abdallah Shami}
\IEEEauthorblockA{\textit{Electrical and Computer Engineering} \\
\textit{Western University}\\
London, Canada \\
Abdallah.Shami@uwo.ca}
}

\IEEEoverridecommandlockouts
\IEEEpubid{\makebox[\columnwidth]{978-1-7281-8416-6/20/\$31.00~\copyright2020 IEEE \hfill} \hspace{\columnsep}\makebox[\columnwidth]{ }}

\maketitle

\begin{abstract}
Convolutional Neural Network (CNN) techniques have proven to be very useful in image-based anomaly detection applications. CNN can be used as deep features extractor where other anomaly detection techniques are applied on these features. For this scenario, using transfer learning is common since pre-trained models provide deep feature representations that are useful for anomaly detection tasks. Consequentially, anomaly can be detected by applying similarly measure between extracted features and a defined model of normality. A key factor in such approaches is the decision threshold used for detecting anomaly. While most of the proposed methods focus on the approach itself, slight attention has been paid to address decision threshold settings. In this paper, we tackle this problem and propose a well-defined method to set the working-point decision threshold that improves detection accuracy. We introduce a transfer learning framework for anomaly detection based on similarity measure with a Model of Normality (MoN) and show that with the proposed threshold settings, a significant performance improvement can be achieved. Moreover, the framework has low complexity with relaxed computational requirements.
\end{abstract}


\IEEEpeerreviewmaketitle

\section{Introduction}
Machine learning-based approaches have been widely used for anomaly and outlier detection in several fields such as data networks \cite{b1, b2, b3, b4, b5, b6, b7, b8}. wireless sensor networks \cite{b9, b10, b11} and manufacturing \cite{b12, b13}. Accurate industrial inspection is one of the main challenges in the manufacturing industry\cite{b14}; An efficient inspection process will help to reduce the overall manufacturing cost while maintain quality requirements. Computer-vision based approaches that use convolutional neural network (CNN) for anomaly detection in manufactured surface textures are commonly used in industrial quality inspection due to their higher accuracy. Besides anomaly detection accuracy, computing requirements in terms of time and memory are other important factors that should be considered as well. Anomaly detection can be generally defined as “the task of recognising that test data differ in some respect from the data that are available during training”\cite{b15}. In CNN-based anomaly detection, this definition assumes that training is conducted using normal images only to build a model that defines normality. This assumption meets practical requirements and matches real-world situations when anomalies samples are usually not available or insufficient to model the anomalous behaviors. Accordingly, anomaly detection techniques can be categorized into the following categories\cite{b15}: Probabilistic approaches, distance-based anomaly detection, reconstruction-based anomaly detection, domain-based detection approaches and information-theoretic techniques. Probabilistic approaches involve estimating the probability density function (pdf) of the training data. Hence, the resultant distribution models the normality, anomaly then can be detected by setting a threshold that defines the normal boundary. Distance-based approaches use distance metrics to measure the similarity between evaluation data and a model of normality. reconstruction-based techniques attempt to reconstruct a copy of the evaluation data according to learned model of normality. Hence, anomaly can be detected by comparing the input with its reconstruction. Domain-based approaches define normal domain by creating boundary of normality based on training data. Thus, evaluation data that fall outside the domain are classified as anomalies. Information-theoretic methods utilize the fact that anomaly presence in the data would alter its information-content and hence, model of normality is defined using computed information-content of training data. Consequentially, anomaly is detected by applying similarly measures between extracted information-contact of evaluation data and normality model. A common theme among these categories is that they attempt to build a Model of Normality (MoN) using useful deep representations extracted from normal training data, anomaly then can be detected by measuring similarity between deep presentations of evaluation data and the MoN. Extracting useful feature presentations from images can be achieved by utilizing transfer learning\cite{16, 17, 18, 19} or by training a CNN from scratch. In contrast to transfer learning, training CNN model from scratch requires availability of training set consists of anomaly-free images. The size of the training set should be large enough to serve training purpose. In many real-world applications, such requirement is not possible or expensive to achieve. This fundamental problem of insufficient training data can be solved by using transfer learning to obtain deep feature representations from models pre-trained on large datasets without constraints on the relation between these training datasets and the target data\cite{17} . Here, knowledge is transferred from the pre-trained model “source domain” to the subject model “target domain”\cite{17}. Transfer learning has been widely used for different anomaly detection applications with very promising results \cite{20, 21, 22, 23, 24, 25}.\\
In this paper, we utilize transfer learning to detect anomalies in images based on similarity measure with a defined Model of Normality (MoN). Our main contributions are: We introduce a low-complexity end-to-end transfer learning framework for anomaly detection based on similarity measure with a Model of Normality (MoN). Further, we address the problem of decision threshold setting and propose a well-defined method to set working-point threshold that leads to significant improvement in detection accuracy. Moreover, we show that transfer learning-based approaches can outperform CNN architectures that are specifically designed and explicitly trained from scratch to achieve the same task.\\
The paper is outlined as follows. The next section provides literature review on related work. Sections 3 and 4 present the anomaly detection framework and experimental setup, respectively. Section 5 discuses the results, the paper is finally concluded in Section 6. 
\begin{figure*}
\centerline{\includegraphics[width=0.8\textwidth]{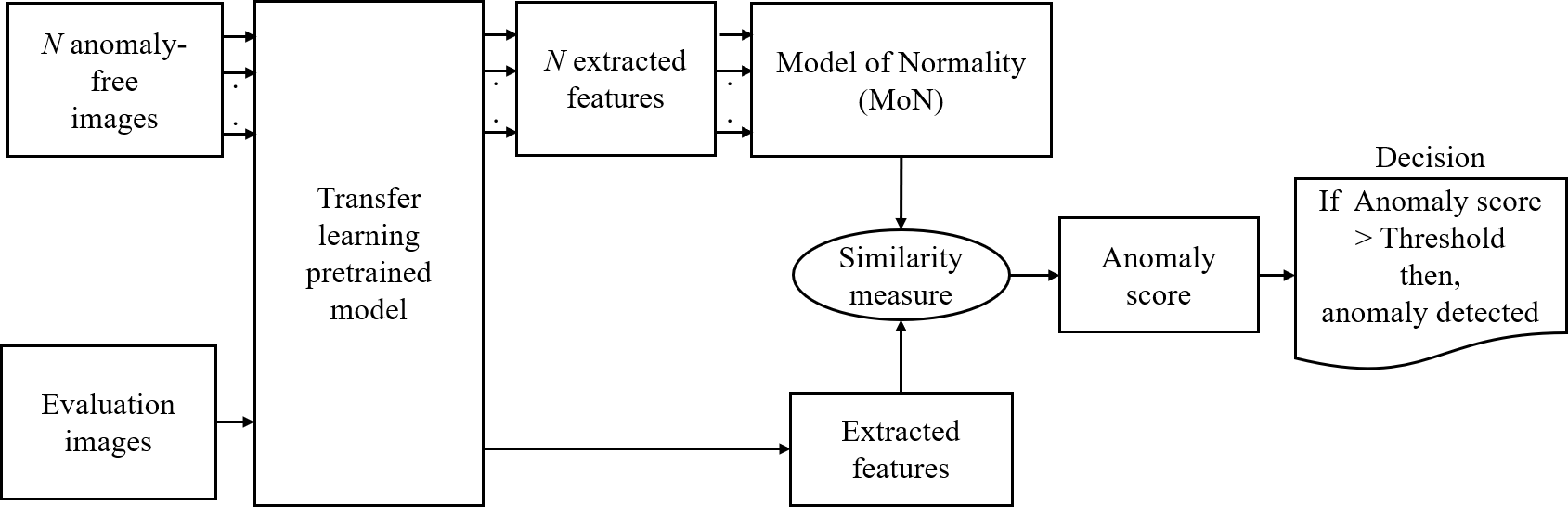}}
\caption{Anomaly detection framework.}
\label{fig}
\end{figure*}

\section{RELATED WORK}
Employing similarity measures with normality model for detecting anomaly in images has been widely adopted in industrial applications due to its effectiveness in detecting anomalous patterns. For example, P. Napoletano \emph{et al.}, (2018) \cite{26} and D. Carrera \emph{et al.}, (2017) \cite{27} use region-based methods to define normality, where a “dictionary” of normality is created from subregions “patches” taken from anomaly-free images. Anomalies subregions of a given image are detected through comparison with normal subregions of the dictionary. Since such methods are region-based, they allow for anomaly localization as well. Other approaches \cite{28, 29, 30, 31, 32, 33, 34, 35, 36} use sparse representations of normal data to learn normality. In O. Rippel \emph{et al.}, (2020) \cite{20} and P. Christiansen \emph{et al.}, (2016) \cite{37}, a Gaussian distribution was fitted to the deep feature representations of normal data to establish the normality model. Subsequently, the Mahalanobis distance \cite{38} is used for anomaly detection. In B. Staar \emph{et al.}, (2018) \cite{39}, a prototype of normal feature representations is created by averaging the learned deep representations of normal images. Here, instead of using normal images directly to define normality, “deep” normality is learned from deep feature representations of these normal images. Euclidean distance was used to measure the similarly between prototype and feature representation of evaluation images. J. Liu \emph{et al.}, (2020) \cite{50} propose an encoder-decoder-encoder CNN structure for anomaly detection in industrial product surface images. They introduced a dual prototype loss approach to encourage features vectors generated by the encoders to keep closer to their own prototype. Consequently, the mean square error between the features vectors is used as indicator of anomalies.

\section{ANOMALY DETECTION FRAMEWORK}
Fig 1. Shows anomaly detection flowchart where $N$ anomaly-free images are fed into transfer learning model (deep features extractor). The learned features are then used to create Model of Normality (MoN) that learns normality from the input $N$ anomaly-free images. For a given image, its features extracted from the transfer learning model are compared to the MoN through a similarity measure and anomaly will be detected if the resultant anomaly score is above decision threshold.
\subsection{Transfer Learning Model}
For transfer learning, we use EfficientNet pretrained on ImageNet dataset \cite{40}. It was introduced by Google Brain Team in 2019 and achieved the top results on ImageNet \cite{40}. EfficientNet uses a new scaling method that uniformly scales
all dimensions (depth, width, and resolution) using a compound scaling coefficient. This sort of balanced scaling of architecture dimensions can lead to better performance \cite{40}. The baseline network of EfficientNet “EfficientNet-B0” was developed by leveraging a multi-objective Neural Architecture Search (NAS) \cite{41} that optimizes both accuracy and Floating-Point Operations Per Second (FLOPS). The baseline network was then scaled using different compound coefficients to obtain the EfficientNet scaled versions “EfficientNet-B1 to B7” \cite{40}.

\subsection{Model of Normality (MoN)}
As mentioned earlier, MoN defines normality according to the deep feature representations of images that used to create it. Once MoN is defined, representations that deviates from MoN (deviation here is defined and controlled by decision threshold setting) will be flagged as anomalies. Hence, for a specific application, it is essential to include all normal variations (or deviations that to be tolerated as anomaly-free) in the MoN creation. In this paper, we use the same method as in [39] to create our model of normality. However, in contrast to \cite{39}, we address model of normality using transfer learning approach rather than training a CNN from scratch. For each class of data, we create a MoN by averaging the learned deep representations of $N$ normal images. These $N$ normal images are used exclusively for MoN creation and are not used as a part of evaluation set.

\subsection{Similarity Measure}
Since each input image is represented by its deep-learned features, the subjective similarity between the MoN and a given image is quantified in terms of a distance measure which is defined on the learned feature space. In the literature, several similarity measures have been proposed to find similarly between images or feature representations such as Structural Similarity Index Measure (SSIM) and Feature Similarity Index Measure (FSIM) \cite{42, 43, 44, 45}. Common similarity measures include the Minkowski distance, the Manhattan distance, the Euclidean distance and the Hausdorff distance \cite{46}. In this work, we use the Euclidean distance to assess the similarity between MoN and deep feature representations of test images.

\begin{table}[b]
\caption{PROPOSED WORKING-POINT THRESHOLDS}
\centering
\begin{tabular}{|c|c|}
\hline
\textbf{Threshold} & \ $D1 = D_{max}, D2 = D_{mean}$ \\    \hline
Threshold1 & $max(D1)$\\    \hline
Threshold2 & $max(D1)-std(D1)$\\    \hline
Threshold3 & $mean(D1)+std(D1)$\\    \hline
Threshold4 & $max(D2)$\\    \hline
Threshold5 & $max(D2)-std(D2)$\\    \hline
Threshold6 & $mean(D2)+std(D2)$\\    \hline
\end{tabular}
\label{table}
\end{table}
\begin{figure*}
\centerline{\includegraphics[width=0.8\linewidth]{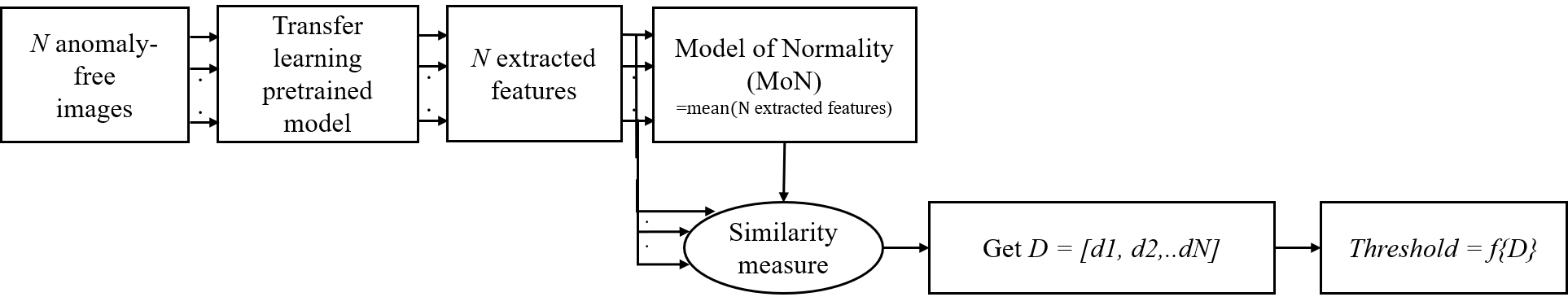}}
\caption{Process of working-point threshold setting.}
\label{fig}
\end{figure*}
\begin{figure*}[htbp]
\centerline{\includegraphics[width=0.8\linewidth]{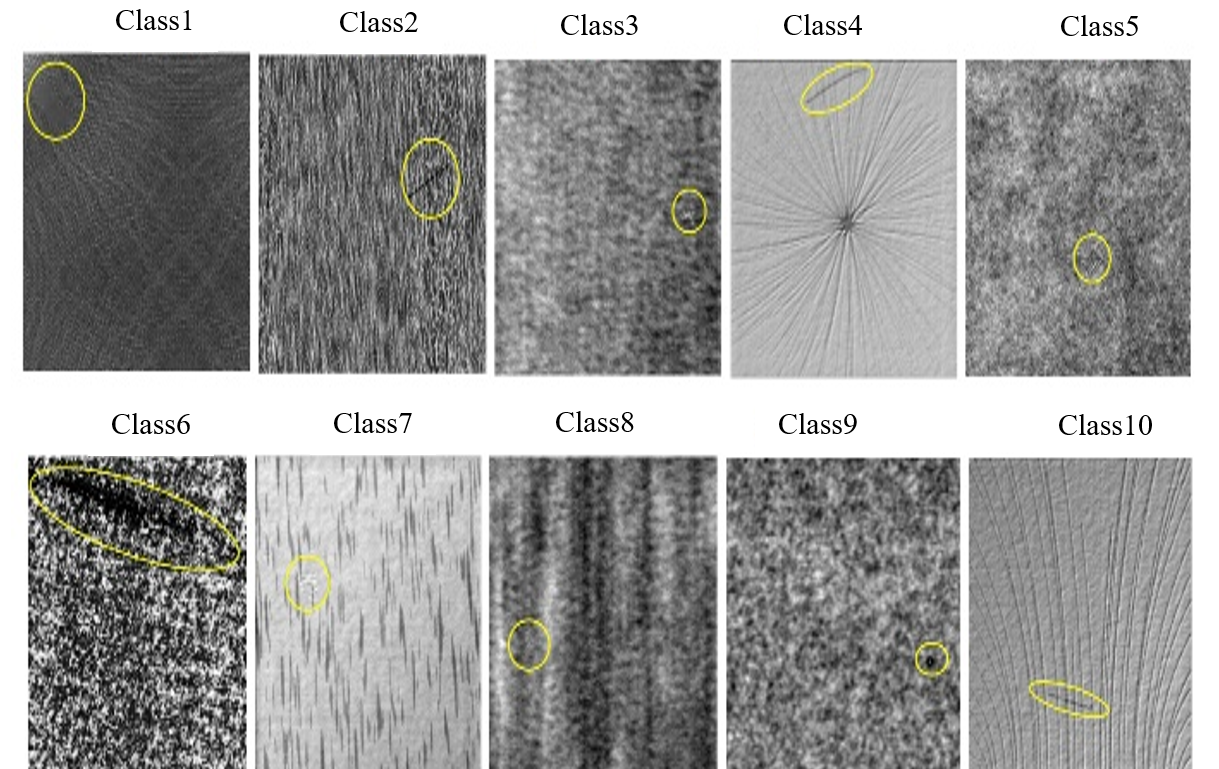}}
\caption{DAGM dataset: defective sample from each class.}
\label{fig}
\end{figure*}
\subsection{Working-Point Decision Threshold}
The decision threshold is a key factor in distance-based anomaly detection methods and has direct impact on detection performance. With the proper setting of this threshold, false positive rate (FPR) will be minimised and detection accuracy can be significantly improved. Most of the proposed methods emphasis on the approach itself and less attention is paid to decision threshold setting. In \cite{39}, varying thresholds were applied the maximum value of the output form Euclidean distance measure between test images and model of normality. \cite{26} uses a region-based method where model of normality is created from subregions taken from \text{anomaly-free} images. Accordingly, the threshold was found by estimating the distribution of the visual self similarity among subregions taken from anomaly-free images. In \cite{27} and \cite{29}, sparse-representations based approaches were used for anomaly detection. Thresholds values were empirically chosen to provide desirable FPR. Here, we propose a well-defined method to set working-point threshold. As mentioned earlier, we create MoN by taking the mean of feature representations of $N$ anomaly free images. Let \text{$T$} represents transformation function of the CNN that transfer the image into its deep feature representations, then \text{$MoN$} can be expressed mathematically as: \\
\begin{equation}
MoN={\frac{1}{N} \sum_{i=1}^{N}T(A_i)}
\end{equation}
Where \text{$Ai, i= 1,2,…N$} are the anomaly-free images. Let \text{$S$} denotes similarity measure operation, then for a given image $B$, the resultant distance array (or distance heatmap) \text{$D_B$} between \text{$B$} and \text{$MoN$} can be expressed as:\\
\begin{equation}
D_B=S(T(B), MoN)
\end{equation}
Since anomaly presence in the image usually results in large distances to the $MoN$, we can describe each image by its mean and maximum distances from the $MoN$, i.e.,\\
\begin{equation}
image=(d_{B_{mean}}, d_{B_{max}})
\end{equation}
where,\\
\begin{equation}
d_{B_{mean}}=mean(D_B),
\end{equation}
\begin{equation}
d_{B_{max}}=max(D_B)
\end{equation}
Thus, anomaly in given image can be detected by setting thresholds on its mean or (and) maximum distances (\text{$d_{B_{mean}}$} and \text{$d_{B_{max}}$}) to the \text{$MoN$}. Working-point threshold can be selected by exploiting deep feature representations of normal images that were used to create \text{$MoN$}. As shown in the flowchart of Fig. 2, for each image \text{$Ai$}, we calculate its distance array \text{$D_{Ai}$}:

\begin{equation}
D_{Ai}=S(T(A_{i}),MoN), i=1,2,...,N
\end{equation}
\begin{equation}
\begin{aligned}
D_{mean}=mean(D_{Ai})=\\
[d1_{mean},d2_{mean},...,dN_{mean}],i=1,2,...,N
\end{aligned}
\end{equation}
\begin{equation}
\begin{aligned}
D_{max}=max(D_{Ai})=\\
[d1_{max},d2_{max},...,dN_{max}],i=1,2,...,N
\end{aligned}
\end{equation}
These distance vectors can be treated as upper pounds for anomaly-free distances to the \text{$MoN$}. Thus, provide useful working-point thresholds for anomaly detection. Table I. shows proposed settings for working-point thresholds based on vectors \text{$D_{max}$} and \text{$D_{mean}$}. Specifically, thresholds1 and thresholds4 equal maximum values of \text{$D_{max}$} and \text{$D_{mean}$}, respectively. However, these thresholds are very sensitive to outliers in \text{$D_{max}$ and $D_{mean}$}, which could impact detection accuracy. This can be solved by introducing marginal versions of these thresholds (Threshold2, Threshold3, Threshold5, Threshold6 in Table II). Here we margined thresholds with a value equals to the standard deviation.

 \section{DATASET AND EXPERIENTIAL SETUP}
For transfer learning model, we employ EffeientNet-B4 were deep feature extracted from stage7 \cite{40}. For an input image width of \text{$224\times224\times3$}, this yields feature representations with dimensions of \text{$14\times14\times192$}. Our choice is based on the work in \cite{20}, it shows that using multivariant Gaussian on features learned from stage7 of medium variants of EfficientNet (B4 and B5) has best performance compared to other variants and stages of EfficientNet.
 \subsection{Dataset}
In this paper, we use DAGM dataset \cite{47} which consists of $512\times512$ grayscale images of statistically textured surfaces. The data is artificially generated but simulates real world problems. Anomalies in the images represent different kinds of defects on varying background surfaces, which makes it suitable and diverse enough to serve the purpose of this paper. The dataset consists of 10 classes, classes 1- 6 consist of 1000 anomaly-free images and 150 anomalies images. Classes 7-10 consist of 2000 anomaly-free images and 300 anomalies images. Fig. 3 shows defective sample from each class.
\subsection{Experiential Setup}
For Classes 1-6, we used 500 anomaly-free images to create MoN. The remaining 500 anomaly-free images with the 150 defective images were used as evaluation set. For classes 7-10, 500 anomaly-free images were used to create the MoN. Another 500 anomaly-free images with the 300 defective images were used as evaluation set. The model was implemented using Python, keras library \cite{48} and TensorFlow \cite{49}.
\begin{figure*}[htbp]
\centerline{\includegraphics[width=\linewidth]{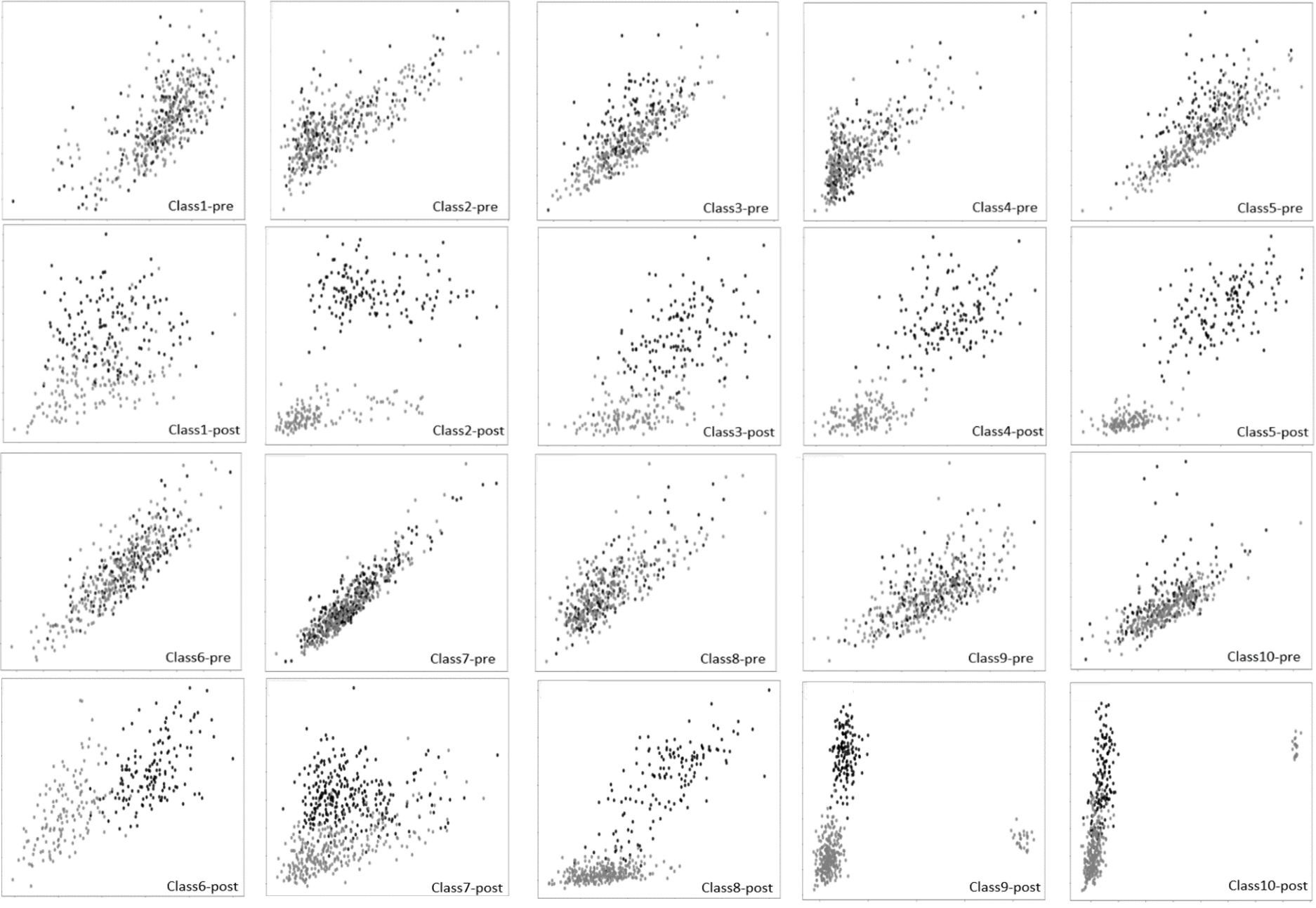}}
\caption{ Pre and post scatter plots of anomaly-free and anomalies images from evaluation set. X-axis is the normalized mean distance while Y-axis is the
normalized maximum distance. Gray and black dots represent anomaly-free images and anomalies images, respectively.}
\label{fig}
\end{figure*}
\begin{table*}[htbp]
\caption{AUC SCORE RESULTS}
\centering
\begin{tabular}{|c|c|c|c|c|c|c|c|c|}
\hline
Class & Threshold1 & Threshold2 & Threshold3 & Threshold4 & Threshold5 & Threshold6 & $max(AUC Score)$ & {Results from [39]}\\
\hline
1 &0.51 &0.55 &0.8 &0.5 &0.5 &0.55 &0.80 &0.98\\
2 &1.00 &1.00 &0.92 &0.57 &0.60 &0.66 &1.00 &0.47\\
3 &0.69 &0.80 &0.92 &0.52 &0.59 &0.76 &0.92 &1.00\\
4 &0.97 &0.99 &0.93 &0.86 &0.94 &0.91 &0.99 &0.49\\
5 &1.00 &1.00 &0.93 &0.88 &0.94 &0.92 &1.00 &1.00\\
6 &0.50 &0.50 &0.67 &0.88 &0.97 &0.92 &0.97 &0.98\\
7 &0.51 &0.54 &0.87 &0.50 &0.50 &0.53 &0.87 &0.95\\
8 &0.93 &0.94 &0.92 &0.69 &0.79 &0.84 &0.94 &1.00\\
9 &0.96 &0.98 &0.92 &0.50 &0.47 &0.47 &0.98 &0.98\\
10 &0.50 &0.54 &0.85 &0.50 &0.47 &0.47 &0.85 &0.57\\
\hline
Mean AUC Score &0.76 &0.78 &0.87 &0.64 &0.68 &0.70 &0.93 &0.83\\
\hline
\end{tabular}
\label{table}
\end{table*}

\section{RESULTS AND DISCUSSION}
Fig.4 shows scatter plots of anomaly-free and anomalies images from evaluation set. Post-scatter plots were generated using deep feature representations of the images according to Equation (3). Pre-scatter plots were generated using images directly by skipping the transfer learning function. As we can see from the plots, deep feature representations extracted by pre-trained transfer learning function are very useful for anomaly detection tasks as they provide good separation between normal and anomalies images in terms of mean and (or) maximum distances with respect to the MoN.\\
Regarding performance results, Table II. shows AUC score results along with comparison with the best results of \cite{39}. As we can see, transfer learning-based framework with the proposed threshold settings outperforms the CNN in \cite{39}, which was built and trained for the purpose of learning similarity metric for surface textures. It is obvious that proposed thresholds perform differently on dataset classes. This is expected sine classes have different anomalies behaviors and hence, anomaly presence in each class alters Euclidean distances to the MoN in a different way. For example, from Fig. 4 post plots, we can see that in classes 2, 4, 5, 8 and 9, normal and anomalies samples are well separated by their maximum distances to the MoN with very few outliers. Thus, Threshold1 and Threshold2 work very well here. On the other hand, for classes, 1, 3, 7 and 10 where there are many outliers, Threshold3 achieves the best results. For class 6, as we can see from its post scatter plot, normal and anomalies samples are better separated by their mean distances to the MoN rather than maximum distances. Thus, Thresholds4, 5 and 6 achieve better separation than other thresholds.\\
Regarding complexity, the proposed framework has low computational
complexity that meets real-time application requirements. Specifically, it requires 0.9985±0.017 (mean±std) seconds with 0.43/3.43(allocated/peak) MB of memory to process one image on a machine with i7-8550u CPU 1.8GHz and 8 GB RAM.

\section{CONCLUSION}
In this paper, we have introduced an end-to-end framework for anomaly detection. The framework achieves the anomaly detection task by utilizing deep feature representations of a transfer leaning model along with similarity measure to model of normality. Thus, it provides a low-complexity solution to implement without the need for training. Further, we proposed a well-defined method to set working-point decision threshold that leads to improved accuracy in anomaly detection. Moreover, we show that using transfer learning-based model with proper threshold settings can outperform CNN that was specifically designed and trained to achieve the same task. The framework has low requirements in terms of computing resources, which makes it suitable for real-time industrial applications.

\section*{Acknowledgment}

This work was funded in part by National Research Council Canada under Project no.: AM-105-1.

\end{document}